\documentclass[conference]{IEEEtran}
\IEEEoverridecommandlockouts
\usepackage{amsmath,amssymb,amsfonts}
\usepackage{algorithmic}
\usepackage{graphicx}
\usepackage{textcomp}
\usepackage{xcolor}

\def\BibTeX{{\rm B\kern-.05em{\sc i\kern-.025em b}\kern-.08em
    T\kern-.1667em\lower.7ex\hbox{E}\kern-.125emX}}

\begin{document}

\title{MonoLite3D: Lightweight 3D Object Properties Estimation\\

}

\author{\IEEEauthorblockN
{1\textsuperscript{st} Ahmed El-Dawy}
\IEEEauthorblockA{
\textit{Electrical Power Department} \\
\textit{Faculty of Engineering} \\
\textit{Alexandria University}\\
Alexandria, Egypt \\
ahmed.dawy@alexu.edu.eg }
\and

\IEEEauthorblockN{
2\textsuperscript{nd} Amr El-Zawawi}
\IEEEauthorblockA
{\textit{Electrical Power Department} \\
\textit{Faculty of Engineering} \\
\textit{Alexandria University}\\
Alexandria, Egypt \\
amr.elzawawi@yahoo.com}
\and

\IEEEauthorblockN{
3\textsuperscript{rd} Mohamed El-Habrouk}
\IEEEauthorblockA{
\textit{Electrical Power Department} \\
\textit{Faculty of Engineering} \\
\textit{Alexandria University}\\
Alexandria, Egypt \\
eepgmme1@yahoo.co.uk}
}

\maketitle

\begin{abstract}Reliable perception of the environment plays a crucial role in enabling efficient self-driving vehicles.\space\space Therefore, the perception system necessitates the acquisition of comprehensive 3D data regarding the surrounding objects within a specific time constrain, including their dimensions, spatial location and orientation.\space\space
Deep learning has gained significant popularity in perception systems, enabling the conversion of image features captured by a camera into meaningful semantic information.\space\space
This research paper introduces MonoLite3D network, an embedded-device friendly lightweight deep learning methodology designed for hardware environments with limited resources.\space \space MonoLite3D network is a cutting-edge technique that focuses on estimating multiple properties of 3D objects, encompassing their dimensions and spatial orientation, solely from monocular images.\space\space This approach is specifically designed to meet the requirements of resource-constrained environments, making it highly suitable for deployment on devices with limited computational capabilities.\newline
The experimental results validate the accuracy and efficiency of the proposed approach on the orientation benchmark of the KITTI dataset.\space\space It achieves an impressive score of 82.27\% on the moderate class and 69.81\% on the hard class, while still meeting the real-time requirements.
\end{abstract}



\section{Introduction}
\label{sec:introduction}
Efficient 3D object properties estimation is a crucial research problem in various fields, including autonomous driving and robot grasping\cite{liu2021ground}.\space\space By accurately estimating the properties of 3D objects, the machines are enabled to interact with their environment more effectively and make informed decisions.\space\space In the context of autonomous driving, estimating the properties of 3D objects, such as their size, shape, and orientation, is essential for tasks like object detection, tracking, and motion planning\cite{calvert2017will}.\space \space This information helps autonomous vehicles understand the surrounding environment, predict object behavior, and make safe driving decisions\cite{sharma2020evaluation}.\newline

The existing methods for detecting 3D objects can primarily be categorized as either relying on LIDAR-based methods or vision-based methods.\cite{li2022survey}.\space\space LIDAR-based methods are precise and effective, but their high cost restricts their application in various industries\cite{wu2020deep}.\space \space
There are two main types of vision-based methods\cite{qian20223d}: monocular algorithms and binocular algorithms.\space\space Vision-based perception systems are commonly utilized because they are affordable and offer a wide range of features.\space\space However, one major drawback of using monocular vision is that it cannot directly determine depth from image data.\space\space This limitation can lead to inaccuracies in estimating the three-dimensional pose of objects in monocular object detection.\space\space On the other hand, binocular vision, which provides more precise depth information compared to monocular vision, comes at a higher cost.\space\space Additionally, binocular vision has a narrower visual field range, which may not be suitable for certain operating conditions\cite{wu2020survey}.\newline This study introduces MonoLite3D network, a lightweight technique based on deep learning that is designed to accurately determine the 3D characteristics of an object using the information obtained from a detected 2D bounding box.\space \space Cutting-edge 2D object detection models like YOLOv7\cite{wang2022yolov7}, Detectron2\cite{abhishek2021detectron2}, BERT\cite{resnick2022causal} have the potential to evolve into 3D object detectors through the training of a compact and efficient feature extractor.\space\space This extractor would be responsible for capturing features from the input data, which can then be employed to predict both the orientation of the 3D bounding box of the object and its dimensions.\newline
Section \ref{sec:related_work} of the paper discusses the existing research on the deep learning approaches used in 3D object detection.\space\space In Section \ref{sec:monolie3d}, the paper introduces MonoLite3D network and highlights its contributions, which are as follows:
\begin{itemize}
  \item The simple design of MonoLite3D network which is composed of cheap operations.
  \item The choice of an efficient, lightweight, embedded device-friendly feature extractor alleviates the computational burden without sacrificing accuracy.
\end{itemize}
Section \ref{sec:experimental_work} of the paper discusses the experimental work of the proposed MonoLite3D network.\space\space It provides information about the implementation details, the dataset used for training and benchmarking, and the training procedures.\space\space This section aims to provide a comprehensive understanding of how the MonoLite3D network was developed and evaluated.\space\space
Experimental results on the KITTI benchmark confirm the effectiveness of the suggested MonoLite3D network architecture, as discussed in Section \ref{sec:results}.\space\space The paper concludes with a summary of findings in the final section.
\section{Related Work}

\begin{figure*}
    \centering
    \includegraphics[width=0.6\textwidth]{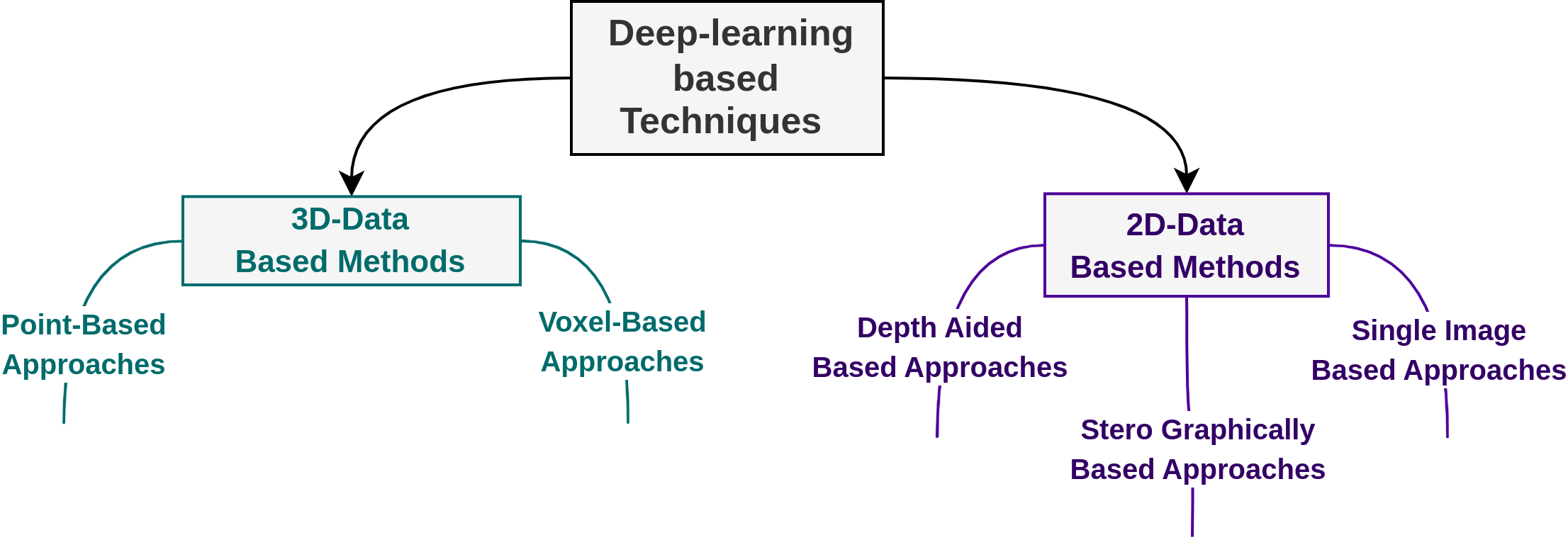}
    \caption{3D object detection deep learning methodologies.}
    \label{fig:related_word}
\end{figure*}

\label{sec:related_work}
This section offers a detailed overview of the current deep learning advancements in 3D object detection.\space\space It focuses on the different methods employed to generate 3D bounding boxes, as depicted in Figure \ref{fig:related_word}.
\subsection{3D-Data Based Methods}
In recent times, there has been notable progress in the field of 3D detection using LIDAR technology \cite{li2021p2v}\cite{li20203d}.\space\space
LIDAR sensors acquire precise 3D measurement data from their surroundings, represented as 3D points with absolute coordinates (x, y, z).\space\space Given the inherent nature of LIDAR point cloud data, it is imperative to employ an architecture that enables efficient convolutional operations.\space\space Consequently, Deep Learning techniques employed for LIDAR-based 3D detection can be categorized into two main approaches: Point-based approaches and Voxel-based approaches \cite{mao20223d}\cite{FERNANDES2021161}.\newline
\textbf{Point-based Approaches:} were developed to handle unprocessed and disorganized point clouds.\space\space Point-based techniques, including PointNet\cite{qi2017pointnet}, utilize raw point clouds as input and extract point-level features for 3D object detection using structures like multi-layer perceptrons.\space\space Other models such as PointRCNN\cite{shi2019pointrcnn}, RoarNet\cite{shin2019roarnet}, and PointPainting\cite{vora2020pointpainting} are instances of deep-learning neural networks that employ the Point-Based approach.\newline
\textbf{Voxel-Based Approaches:} divide point clouds into equally sized 3D voxels.\space\space Then, for each voxel, features can be extracted from a group of points.\space\space This approach reduces the overall size of the point cloud and saves storage space\cite{FERNANDES2021161}.\space\space Voxel-based techniques, such as VoxelNet\cite{zhou2018voxelnet}, extend the 2D image representation to 3D space by dividing it into voxels.\space\space Other deep learning models, Center-based 3D object detection and tracking\cite{yin2021center} and Afdet\cite{ge2020afdet} utilize the Voxel-based approach for solving 3D object detection.

\subsection{2D-Data Based Methods}
Deep-learning techniques that depend on 2D data primarily use RGB images as their main input.\space\space While 2D object detection networks have shown exceptional performance, the task of generating 3D bounding boxes solely from the 2D image plane is considerably more complex due to the absence of absolute depth information\cite{rahman2019notice}.\space\space As a result, approaches for 3D object detection using 2D data can be categorized based on how they address the challenge of obtaining depth information.\space\space These categories include Depth-Aided Based Based Approaches, Stereo-Graphically Based Approaches, and Single Image Based Approaches.\newline
\textbf{Depth-Aided Based Approaches:} As a result of the absence of depth information in single monocular images, several studies have attempted to leverage advancements in depth estimation neural networks.\space\space In earlier studies \cite{ma2019accurate}\cite{wang2019pseudo}, images are transformed into pseudo-LIDAR representations by utilizing readily available depth map predictors and calibration parameters.\space\space Subsequently, established LIDAR-based 3D detection methods are applied to generate 3D bounding boxes, albeit with reduced effectiveness.\space\space
In contrast, DDMP-3D \cite{wang2021depth} emphasize a fusion-based approach that cleverly integrates information from both images and estimated depth through specially designed deep convolutional neural networks (CNNs).\space\space However, most of the previously mentioned methods that directly rely on off-the-shelf depth estimators incur significant computational overhead and yield only marginal improvements due to the inherent inaccuracies in the estimated depth maps\cite{huang2022monodtr}.\newline
\textbf{Stereo-Graphically Based Approaches:}  Approaches outlined in \cite{wang2019anytime}\cite{kendall2017end}\cite{zhang2019ga} involve the processing of a pair of stereo images using a Siamese network.\space\space These methods then generate a 3D cost volume, which is utilized to determine the matching cost for stereo matching through neural networks.\space\space Additionally, MVS-Machine\cite{kar2017learning} adopts a differentiable approach involving projection and re-projection to enhance the construction of a 3D volume from multi-view images.\newline
\textbf{Single Image Based Approaches:}
In recent studies \cite{simonelli2020towards}\cite{zhang2021objects}, the focus has shifted towards using only a single monocular RGB image as input for 3D object detection.\space\space
For instance, PGD-FCOS3D\cite{PGD} introduces geometric correlation graphs among detected objects and leverages these constructed graphs to enhance depth estimation accuracy.\space\space Other approaches such as RTM3D\cite{li2020rtm3d} and SMOKE\cite{liu2020smoke} anticipate key points of the 3D bounding box as a complementary step to establish spatial information about observed objects.\space\space MonoCInIS\cite{heylen2021monocinis} proposes a method that utilizes instance segmentation to estimate an object's pose, and it is designed to be camera-independent to account for different camera perspectives.\space\space Many recent studies build upon a prior 2D object detection stage.\space\space For example, Deep3Dbox\cite{mousavian20173d} presents an innovative approach for predicting object orientation and dimensions.\space\space M3D-RPN\cite{brazil2019m3d} incorporates depth-aware convolution to anticipate 3D objects and generate 3D object properties while satisfying 2D detection requirements.\space\space Consequently, PoseCNN\cite{xiang2017posecnn} identifies an object's position in a 2D image while simultaneously predicting its depth to determine its 3D position.\space\space It's worth noting that estimating 3D rotation directly with PoseCNN is challenging because the rotation space is nonlinear, as pointed out in\cite{wu2020survey}.
\section{MonoLite3D Network}
\label{sec:monolie3d}


By making the assumption that the perspective projection of a 3D bounding box should closely align with its 2D detection window, it becomes feasible to leverage the achievements of previous research in the field of 2D object detection for the estimation of 3D bounding boxes, as demonstrated in earlier work such as\cite{huang2019perspectivenet}.\space\space
The 3D bounding box can be defined by its dimensional attributes D $ = [d_x, d_y, d_z]$, central coordinates T $= [t_x, t_y, t_z]^T$, and orientation R $(\theta ,\phi,\alpha )$, where these parameters are defined by azimuth ($\theta$), elevation ($\phi$), and roll ($\alpha$) angles. In scenarios where the ground is assumed to be flat, it is safe to assume that $\phi$ and $\alpha$ angles are both zero. Furthermore, when all objects are considered to be on the ground, it is reasonable to set the object's height ($t_y$) to zero.\newline
Subsequently, The projection of a 3D point from the object's coordinate frame, denoted as $X_o =[X, Y, Z, 1]^T$, to the image frame, represented as $x = [u, v, 1]^T$, is given by the following equation as detailed in\cite{mousavian20173d},\cite{daniilidis2006imaging},\cite{gu2015camera}:
\begin{align}
x &= K \left[\begin{array}{ccc}
R & T \end{array}
\right]X_o &
\end{align}
This transformation is performed based on the object's pose within the camera coordinate frame $(R, T)\in SE(3)$ and the intrinsic matrix of the camera, denoted as K.\newline
\begin{figure}[!htb]
\centering
\includegraphics[width=0.5\textwidth]
{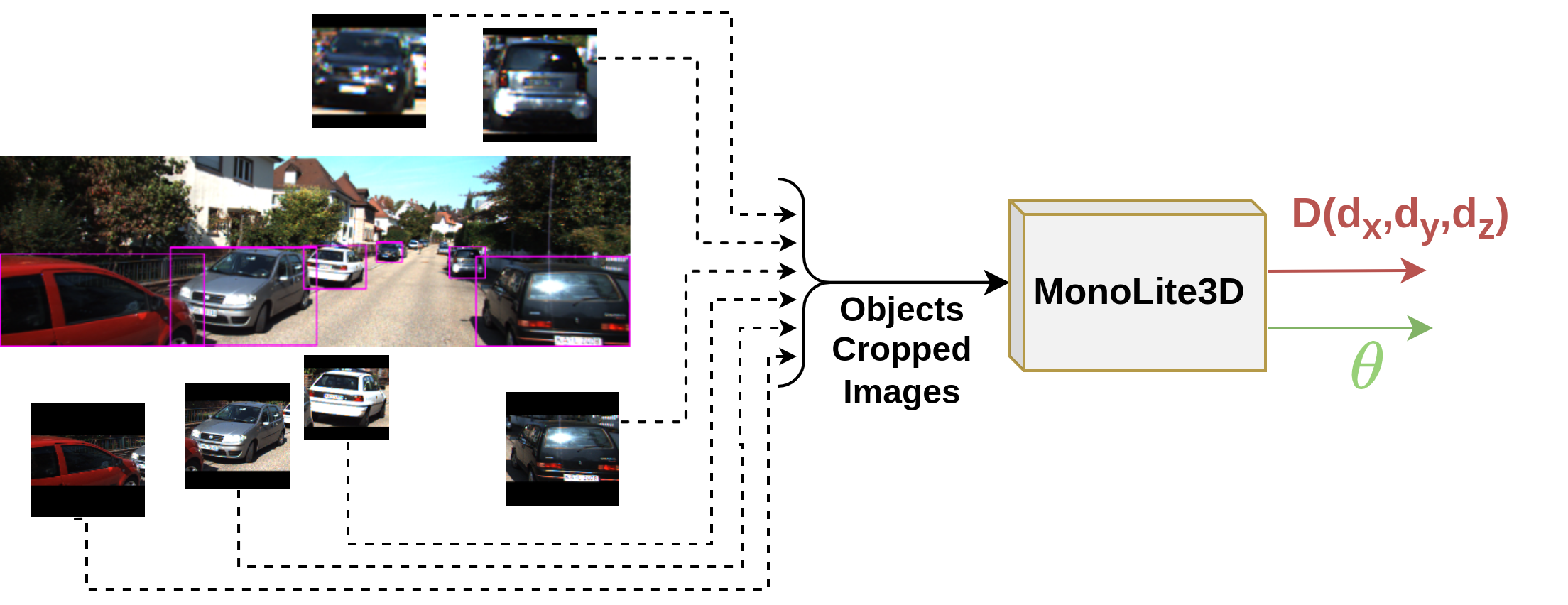}
\caption{The general architecture of the proposed MonoLite3D network is composed of one stage, which is the Orientation-Dimensions Estimator stage.}
\label{fig:general_diagram}
\end{figure}

While projecting a 3D point from the object's coordinate frame onto the image plane is straightforward, it results in the loss of depth information.\space\space Conversely, projecting a point from the image back into the 3D object's coordinate frame is a complex task.\space\space To recover 3D object coordinates from a 2D image, comprehensive feature estimation is vital. That's why the MonoLite3D network prioritizes object orientation and dimensions.\space\space Consequently, MonoLite3D network is composed of one main stage as presented in Figure \ref{fig:general_diagram} :
\begin{itemize}
  \item The Orientation-Dimensions Estimator stage takes the cropped object image as input, proceeds to extract visual characteristics of the object, and subsequently generates the object's geometric attributes, including orientation and dimensions.
\end{itemize}

\subsection{Orientation Estimation}
\begin{figure}[!htb]
\centering
\includegraphics[width=0.5\textwidth]
{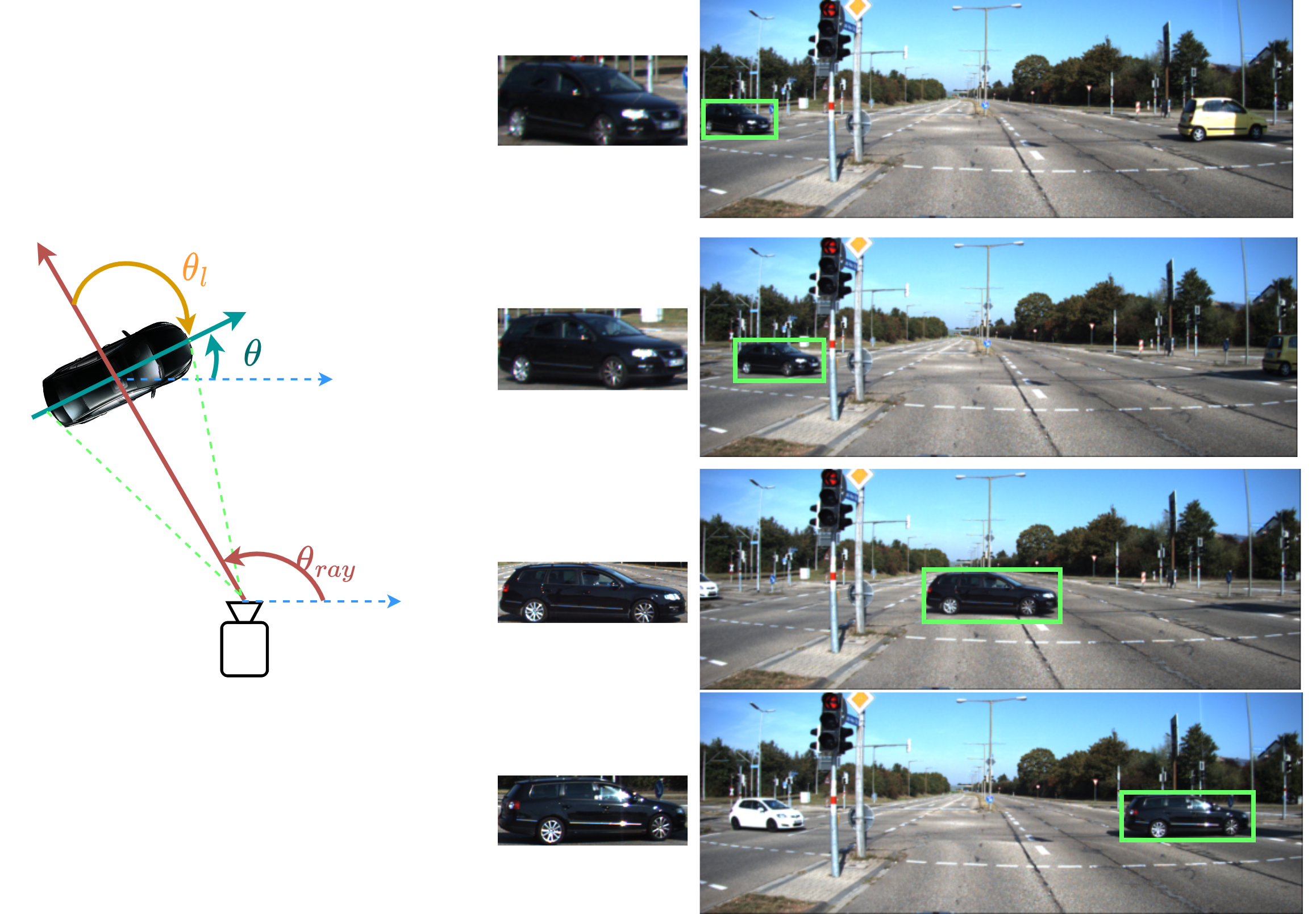}
\caption{The object's orientation, denoted as $\theta$, is computed by adding $\theta_{ray}$ and $\theta_l$ together. The Orientation-Dimensions Estimator provides the value for $\theta_l$, while $\theta_{ray}$ can be determined in relation to the center of the object's bounding box using the known camera intrinsic parameters.}
    \label{fig:theta_local}
\end{figure}

\begin{figure*}[!htb]
    \centering
    \includegraphics[width=0.6\textwidth]{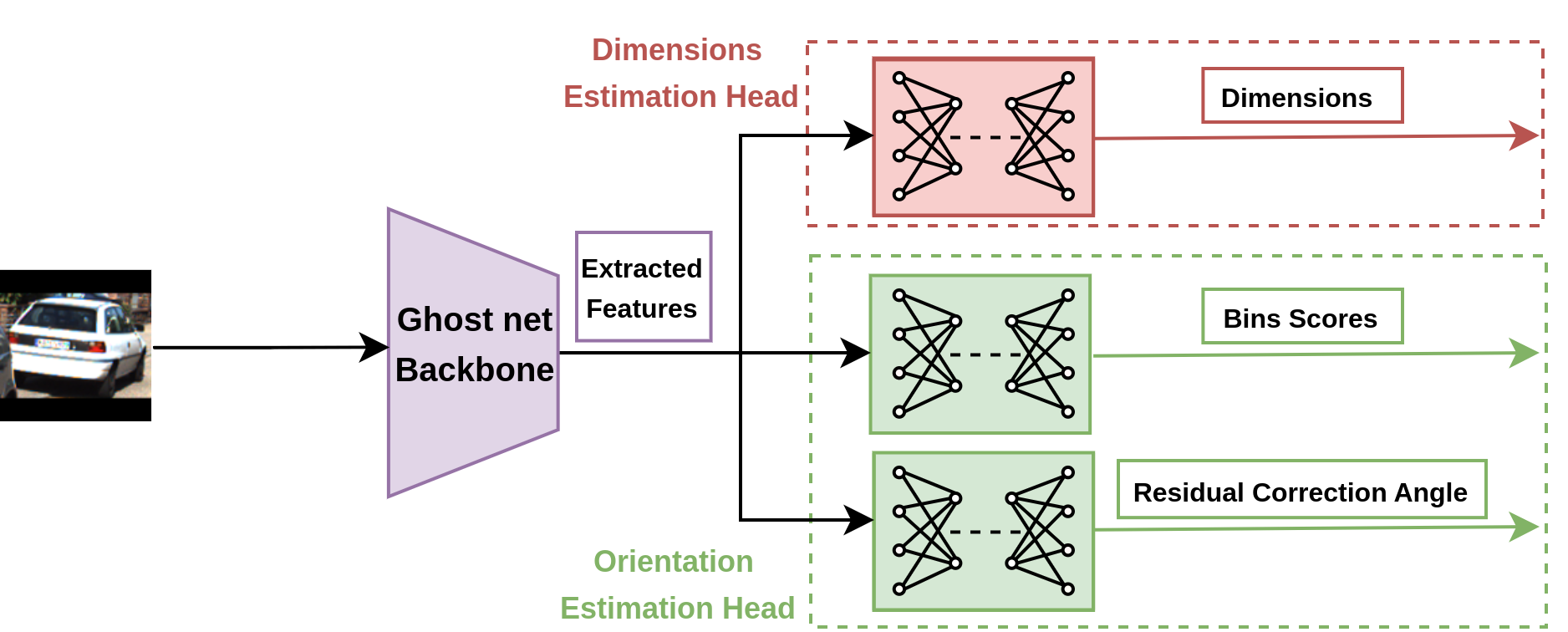}
    \caption{Orientation-Dimensions Estimation heads.}
    \label{fig:orient_dimen}
\end{figure*}
Estimating the object's overall orientation, denoted as $R\in SO(3)$, requires more than just the 2D bounding box information.\space\space It also depends on the box's position in the image plane.\space\space For instance, consider the rotation $R(\theta)$, where $\theta$ (yaw) represents azimuth, as shown in Figure \ref{fig:theta_local}.\space\space Consider the example of a vehicle moving in a straight line.\space\space Here, the global orientation $R(\theta)$ remains constant, but the local orientation $\theta_l$ based on the vehicle's appearance within the 2D bounding box, changes.\space\space Determining the overall direction R($\theta$) involves combining the variation in the local direction $\theta_l$ relative to the ray passing through the bounding box's center.\space\space This process is aided by computing the ray's direction at a specific pixel using intrinsic camera parameters $\theta_{ray}$.\space\space To summarize, the steps include predicting local orientation from 2D bounding box features ($\theta_l$), then combining this with the ray's direction ($\theta_{ray}$) toward the bounding box's center to calculate the object's global orientation.\newline
Object detectors like Yolo and SSD\cite{liu2016ssd} employ anchor boxes to define potential bounding box modes and compute adjustments for each anchor box.\space\space Similarly, MultiBin architecture\cite{mousavian20173d} adopts a related concept for orientation estimation.\space\space It discretizes the angle space into overlapping bins, where each bin is associated with score probabilities ($score_i$) for the angle falling within it and residual correction angles to align with the center ray of the bin.\space\space These corrections are represented by sine and cosine values, yielding three outputs per bin ($score_i$, $sine(\Delta \theta_i)$, $cosine(\Delta \theta_i)$).\newline 
Consequently, the total loss for the MultiBin orientation is:
\begin{align}
L_{\theta} &= L_{score} + \alpha\times L_{residual}&
\end{align}
The softmax loss \cite{kingsbury2009lattice} computes the score loss $L_{score}$ for each bin.\space\space The residual loss $L_{residual}$ minimizes the difference between the estimated angle and the ground truth angle within the relevant bin.\space\space Thus, $L_{residual}$ maximizes cosine distance and is calculated as:
\begin{align}
L_{residual}&=  \frac{1}{n_{\theta ^ *}} \sum cos(\theta ^ *- c_i -\Delta \theta_i)&
\end{align}
where $n_{\theta ^ *}$ is the number of bins spanning the ground truth angle ${\theta ^ *}$.\space \space $c_i$ is the angle of the bin's $i$ centre.\space \space $\Delta \theta_i$ is the adjustment that must be made to the centre of the bin $i$.\newline
As depicted in Figure \ref{fig:orient_dimen}, the Orientation-Dimensions estimator employs Ghostnet \cite{han2020ghostnet} as its backbone for extracting meaningful visual object features.\space\space Subsequently, these extracted features are shared across three distinct fully connected branches.\space\space The first branch consists of (1280, 256, 3) fully connected units and is responsible for estimating object dimensions.\space\space In contrast, the second branch comprises (1280, 256, 2) fully connected units to produce bin scores.\space\space Lastly, the third branch is composed of (1280, 256, 2x2) fully connected units, generating residual correction angles in both sine and cosine representations.

\subsection{Dimensions Estimation}
The KITTI dataset\cite{geiger2012we} includes distinct categories such as cars, vans, trucks, pedestrians, cyclists, and buses, each of which exhibits noticeable similarities in terms of shape and size among objects.\space\space For each dimension, it is convenient to predict the deviation value from the mean parameter value computed across the training dataset.\space\space The loss for dimension estimation can be computed as follows \cite{mousavian20173d}:
\begin{align}
L_{Dimensions}&= \frac{1}{n} \sum (D^*- \Bar{D} - \delta)^2&
\end{align}
Here, $D ^ *$ represents the true dimensions of 3D bounding boxes, $\Bar{D}$ denotes the mean dimensions of objects within a specific class, $n$ stands for the number of objects in the training batch, and $\delta$ represents the predicted deviation value relative to the average value predicted by the neural network.\newline

\subsection{GhostNet As An Off The shelf Lightweight Feature Extractor}
Feature maps are spatial representations derived from applying convolution layers to input data.\space\space These maps capture specific image characteristics based on learned filter weights.\space\space Within standard convolution layers, it's observed that numerous similar intrinsic features exist across all feature maps, termed "Ghost Feature Maps" \cite{han2020ghostnet}.\space\space GhostNet's core aim is to reduce parameters and FLOPs while maintaining performance close to the original feature maps.\space\space It accomplishes this by generating some output feature maps and using a low-cost linear operation for the rest, resulting in fewer parameters and FLOPs.\space\space This operation adapts to input data and supports optimization through backpropagation \cite{han2020ghostnet}.\space\space GhostNet's effectiveness as a feature extractor is evident in various applications \cite{ting2021ship, zhou2023light, chi2023ghostnet}, particularly in tasks like the Orientation-Dimensions Estimator.\space\space This efficient design allows the MonoLite3D network to have a combined parameter count of only 5.61 million while preserving estimation accuracy.
\section{ Experimental Work}
\label{sec:experimental_work}
\subsection{Implementation Details}
As discussed earlier, the proposed MonoLite3D network extracts an object's 3D properties from a monocular image's 2D bounding box.\space\space This is achieved through a key stage, depicted in Figure \ref{fig:orient_dimen}, which includes a feature extractor and three interconnected branches sharing extracted features. Its main function is to estimate the object's 3D dimensions and orientation.
\subsection{Dataset}
The KITTI dataset, introduced in \cite{geiger2012we}, provides a widely accessible open-source resource for evaluating learning-based approaches in realistic autonomous driving scenarios.\space\space Performance evaluation is categorized based on factors like occlusion, truncation level, and the visible height of object instances' 2D bounding boxes, resulting in easy, moderate, and hard test cases as detailed in \cite{geiger2012we}.\space\space The dataset also includes benchmarks like 2D Object Detection and Orientation, used to evaluate the MonoLite3D network's orientation estimation performance.
\subsubsection{Data Augmentation}
To enhance the training dataset, enhance the training procedure, and mitigate overfitting, various data augmentation methods were applied.\space\space These techniques encompass the introduction of random Gaussian noise, optical distortion (with a 0.5 probability), and random fog (with an 0.8 probability).
\subsubsection{Preprocessing}
\begin{figure}[!htb]
\centering
\includegraphics [width=0.2\textwidth]
{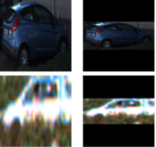}
\caption{The general architecture of the proposed MonoLite3D network is composed of one stage, which is the Orientation-Dimensions Estimator stage.}
\label{fig:resize_padding}
\end{figure}
Complex preprocessing techniques were deliberately avoided when handling the dataset.\space\space Instead, the sole preprocessing operation involved applying resizing with padding to the cropped 2D image of the detected object.\space\space This step's purpose is to retain the object's inherent characteristics before passing it on to either the Orientation-Dimensions Estimator or the Bird's Eye View Center Estimator.\space\space This approach ensures that object properties are preserved, rather than risking the loss of these properties through scaling alone, as depicted in Figure \ref{fig:resize_padding}.
\subsection{Training}
The training of the MonoGhost network was conducted using a Geforce RTX 3060 Ti 8G graphics card.\space\space The choice of optimizer for this process was AdamW, as detailed in \cite{loshchilov2017decoupled}, with a weight decay parameter of 1e-3.
\subsubsection{Orientation-Dimensions Estimator}
The Orientation-Dimensions Estimator was trained with a batch size of 200 for a total of 250 epochs.\space\space Initially, the learning rate was set to 1e-4.\space\space The training employed a scheduler configured to reduce the learning rate on a plateau, with a reduction factor of 0.1, a patience of 10 epochs, and a threshold of 1e-4.\space\space The chosen optimizer for this task was AdamW, incorporating a weight decay of 1e-3.
\section{Results And Discussions}
\label{sec:results}

Given that the MonoLite3D network, as presented, consists of a single primary stage, its performance can be evaluated using the KITTI object orientation benchmark.\space\space To facilitate this benchmark, an off-the-shelf 2D object detector was employed, Faster RCNN\cite{girshick2015fast}, to provide the coordinates of the 2D bounding boxes.\space\space

\indent  TABLEI shows the proposed MonoLite3D network results for orientation on KITTI benchmark.\newline

\begin{table}
\begin{center}
\caption{MonoLite3D network orientation score on KITTI benchmark.}
\begin{tabular}{c| c | c | c}
{\bf Benchmark} & {\bf Easy} & {\bf Moderate} & {\bf Hard}\\ \hline
Car (Detection) & 90.79 \% & 83.33 \% & 71.13 \%\\
Car (Orientation) & 90.23 \% & 82.27 \% & 69.81 \%\\
\hline
\end{tabular}
\end{center}
\label{table:TableI}
\end{table}

\begin{table}
\begin{center}
\caption{Comparison of MonoLite3D network orientation scores on the "Car" class within the KITTI benchmark, across different difficulty levels: Easy, Moderate, and Hard.\space\space The table presents MonoLite3D's orientation performance alongside other state-of-the-art models, highlighting its effectiveness in estimating orientation for car objects.}

\begin{tabular}{c| c | c | c |c}
{\bf Model} & {\bf Easy} & {\bf Moderate} & {\bf Hard}&{\bf Inference Time}\\ \hline
CMAN\cite{CMAN2022}&	89.43 \% &	81.96 \% &	63.74 \% & 0.15 s\\
D4LCN\cite{ding2020learning}& 90.01 \% &	82.08 \% &	63.98 \% &0.2 s\\
Pseudo-LiDAR++\cite{you2020pseudo}&94.14 \% &	81.87 \% &	74.29 \% &0.4 s\\
Disp R-CNN\cite{sun2020disprcnn}&	93.49 \% &	81.96 \% &	67.35 \% & 0.387 s\\
\hline
MonoLite3D & 90.23 \% & 82.27 \% & 69.81 \% & 0.01514 s\\
\hline
\end{tabular}
\end{center}
\label{table:TableII}
\end{table}


The MonoLite3D network's main contributions are its efficient design, based on lightweight operations, and the use of a lightweight feature extractor.\space\space Evaluating its real-time performance is essential.\space\space On a Geforce RTX 3060 Ti 8G, it achieves an average inference time of 0.0121 seconds for a batch of 200 objects, while on a GeForce GTX 1050 Ti, it averages 0.01514 seconds.\space\space These results confirm its real-time capability, even on less powerful GPUs.\space\space MonoLite3D network highlights that using the MultiBin discrete-continuous approach for orientation estimation, in combination with a lightweight feature extractor, significantly improves performance.\newline

The remarkable performance of MonoLite3D extends beyond its accuracy, showcasing efficiency on limited hardware resources.\space\space As illustrated in TABLEII, across difficulty levels on the KITTI benchmark, particularly in Moderate (82.27\%) and Hard (69.81\%) cases, MonoLite3D outperforms competing models, including CMAN\cite{CMAN2022}, D4LCN\cite{ding2020learning}, Pseudo-LiDAR++\cite{you2020pseudo}, and Disp R-CNN. Notably\cite{sun2020disprcnn}, MonoLite3D achieves these commendable orientation scores within an impressive execution time of just 0.01514 seconds on the cost-effective GeForce GTX 1050 Ti, significantly outpacing the inference times of other models such as CMAN\cite{CMAN2022} (0.15 s),  D4LCN\cite{ding2020learning} (0.2 s), Pseudo-LiDAR++\cite{you2020pseudo} (0.4 s), and Disp R-CNN\cite{sun2020disprcnn}. Notably (0.387 s).\space\space This not only positions MonoLite3D as a leader in accuracy but also underscores its unparalleled real-time capabilities.\space\space The model's superior performance on both accuracy and efficiency fronts makes it an efficient and practical choice for real-world applications, ensuring precise orientation estimation while operating on resource-constrained hardware.

\section{Conclusion}
\label{sec:conclusion}
This paper introduced an innovative and lightweight architectural solution for the estimation of complete 3D geometric characteristics of known object classes from a single image.\space\space The MonoLite3D network, designed for monocular 3D geometric feature estimation, demonstrates promising levels of accuracy. It achieves results of 90.23\%, 82.27\%, and 69.81\% on the KITTI object orientation benchmark, all while maintaining an average inference time of just 0.033 seconds per batch of 70 objects.\space\space This proposed network exhibits the potential for seamless integration with state-of-the-art 2D object detection platforms, making it a viable choice for deployment in autonomous vehicles and robotic navigation systems.

\end{document}